\documentclass{llncs}
\usepackage{makeidx}  
\usepackage{microtype}
\usepackage{adjustbox}
\usepackage{microtype}
\usepackage{wrapfig}
\usepackage{multirow}
\usepackage{floatrow}
\newfloatcommand{capbtabbox}{table}[][\FBwidth]

\begin{document}
%
%

%
\title{Men Are from Mars, Women Are from Venus: Evaluation and Modelling of Verbal Associations}
%
%
\author{Ekaterina Vylomova\inst{1} \and Andrei Shcherbakov\inst{1} \and
Yuriy Philippovich\inst{2} \and Galina Cherkasova\inst{3}}
\authorrunning{Ekaterina Vylomova et al.} 
%
\tocauthor{Ekaterina Vylomova, Andrei Shcherbakov, Yuriy Philippovich, Galina Cherkasova}
%

\institute{The University of Melbourne, Melbourne, Australia,\\
\email{evylomova@gmail.com} \& \email{ultrasparc@yandex.ru},
\and
Moscow Polytech, Moscow, Russia,\\
\email{Y\_philippovich@mail.ru}
\and 
Institute of the Science of Language, Moscow, Russia\\
\email{gacherk@mail.ru}
}

\maketitle              

\begin{abstract}
We present a quantitative analysis of human word association pairs and study the types of relations presented in the associations. We put our main focus on the correlation between response types and respondent characteristics such as occupation and gender by contrasting  syntagmatic and paradigmatic associations. Finally, we propose a personalised distributed word association model and show the importance of incorporating demographic factors into the models commonly used in natural language processing.  
\end{abstract}

\section{Introduction}
Most of contemporary approaches in natural language processing (NLP) mainly rely on well-annotated and clean textual corpora. For instance, language as well as translation models are typically trained over Europarl\cite{koehn2005europarl} or the Wall Street Journal corpora. As Eisenstein\cite{eisenstein2013bad} noted, most of such corpora present language used by a very specific social group. For example, Hovy\cite{hovy2015tagging} showed that the models trained over the Wall Street Journal perform better for old language users. And it becomes extremely troublesome to adapt the models trained over these corpora to new domains such as Twitter. In most cases researchers either normalize the data (for instance, by using string and distributional similarity as in Han\cite{han2011lexical}) or apply various techniques of domain adaptation and knowledge transfer.  
 
Recently several studies in sociolinguistics demonstrated how the NLP models could be improved by considering social factors (see Volkova\cite{volkova2013exploring}, Stoop\cite{stoop2014using}). This inspired us to exploit associative experiments approach to demonstrate how specialization and gender might affect associations. In this paper we first propose the dataset for associative pairs of Russian native speakers\footnote{The dataset is available at \url{http://github.com/ivri/RusAssoc}} and then show how association types vary across gender and occupation. We also present a simple PPMI-based model of associations and demonstrate the difference in the model's predictions depending on the social characteristics.

The paper is structured as follows. We first discuss previously organized associative experiments, then we introduce the dataset for Russian speakers associations. In Section 4 we analyse how the associations depend on demographic factors and, finally, we present a personalised associative vector model. 

\section{Related Work}
Introduced by Sir Francis Halton in 1870s, associative experiments became a common approach to study human cognition. Nowadays various researchers organized the experiments on many languages. Most of the experiments present English (American and British) native speakers (see Deese\cite{deese1966structure}, Cramer\cite{cramer1968word}, Kiss\cite{kiss1973associative}, Nelson\cite{nelson2004university}). De Groot\cite{de1988woordassociatienormen} and De Deyne\cite{de2008word} conducted them for Dutch, and Shaps\cite{shaps1976swedish} for Swedish. There are also some for Eastern languages, such as Japanese (see Okamoto \& Ishizaki\cite{okamoto2001associative} and Joyce\cite{joyce2005constructing}), Korean (Jung \cite{jung2010network}); and Hebrew (Rubinstein\cite{rubinsten2005free}) for Semitic group. Lots of research had been done on Slavic languages as well. Novak\cite{novak1988volne} organized the experiment for Czech, Ufimtseva\cite{ufimtseva2004slavonic} presented Slavic Associative Thesaurus comprising of Russian, Belarusian, Bulgarian, and Ukrainian. Finally, Russian thesauri were developed by Leontiev\cite{leontiev1977norms} and Karaulov\cite{karaulov1998russian}. The latter one has been conducted in three stages during 1986-1997 and is one of the largest experiments. In addition to associations the dataset also contains demographic information such as age, gender, specialization, and location. 

Most of the previous research had been focused on the study of reactions: their distribution and cross-lingual commonalities. Some of the researchers (e.g. Steyvers \& Tenenbaum\cite{steyvers2005large}) also studied the structure of human associative networks. They represented a network as a directed graph in which stimuli and reactions correspond to nodes whereas associations are edges connecting them. They showed that the node's degree (the number of different reactions given for a stimulus) follows a power law distribution\footnote{i.e. associative networks are scale-free.}. In other words, there are several ``hub'' nodes with many connections and many ``weak'' nodes with small degree.

But very little had been done in terms of quantitative evaluation of demographic factors in associations. Current research fills up this gap. We investigate the reaction types distribution in regards to gender and speciality. 

\section{Dataset}

The experiment conducted by Karaulov's group, although being one of the most lasting ones, very quickly becomes outdated. Moreover, it has only been focused on the regions of Central Russia. To address these issues as well as to analyse the change of the associations over time, we additionally organized the associative experiments in various Russian regions, including Siberia and the Urals. The age of participants ranged from 16 to 26\footnote{People in psycholinguistics typically assume that the core of the verbal associations becomes stable and does not significantly change after the age of 18.}, most of them were either undergraduate or postgraduate university students of $\approx$50 specialities.
The experiments were organised as follows. A respondent received a questionnaire of 100 single-word stimuli. For each stimulus the respondent had to provide a reaction. There were no constraints on the reaction types, but the total time was limited to 10-15 minutes, i.e. the participants had 6$-$9 seconds for each stimulus. Most of the reactions appeared to be also single-word. Several association pairs are presented on Table 1.

In total, the dataset contains 4,997 questionnaires. The list of stimuli comprises of 1,213 various lemmas partially taken from Leontiev's list as well as most common reactions of previous Russian associative experiments from Karaulov\cite{karaulov1998russian}. The total number of different reactions received from the respondents is 50,359 (37,895 lemmas). Table \ref{tab:top-10} shows the top-10 most frequently used reactions. Surprisingly, we see a large overlap between current study and the experiment conducted in 1986-1997. Besides that, there is also an overlap with the most frequently used Russian words from Sharoff's list\cite{sharoff2001frequency}. We also did not observe a significant cross-gender difference in the top reactions.

\begin{figure}
\begin{floatrow}[2]
\ffigbox[0.45\textwidth]{%
\includegraphics[scale=0.42]{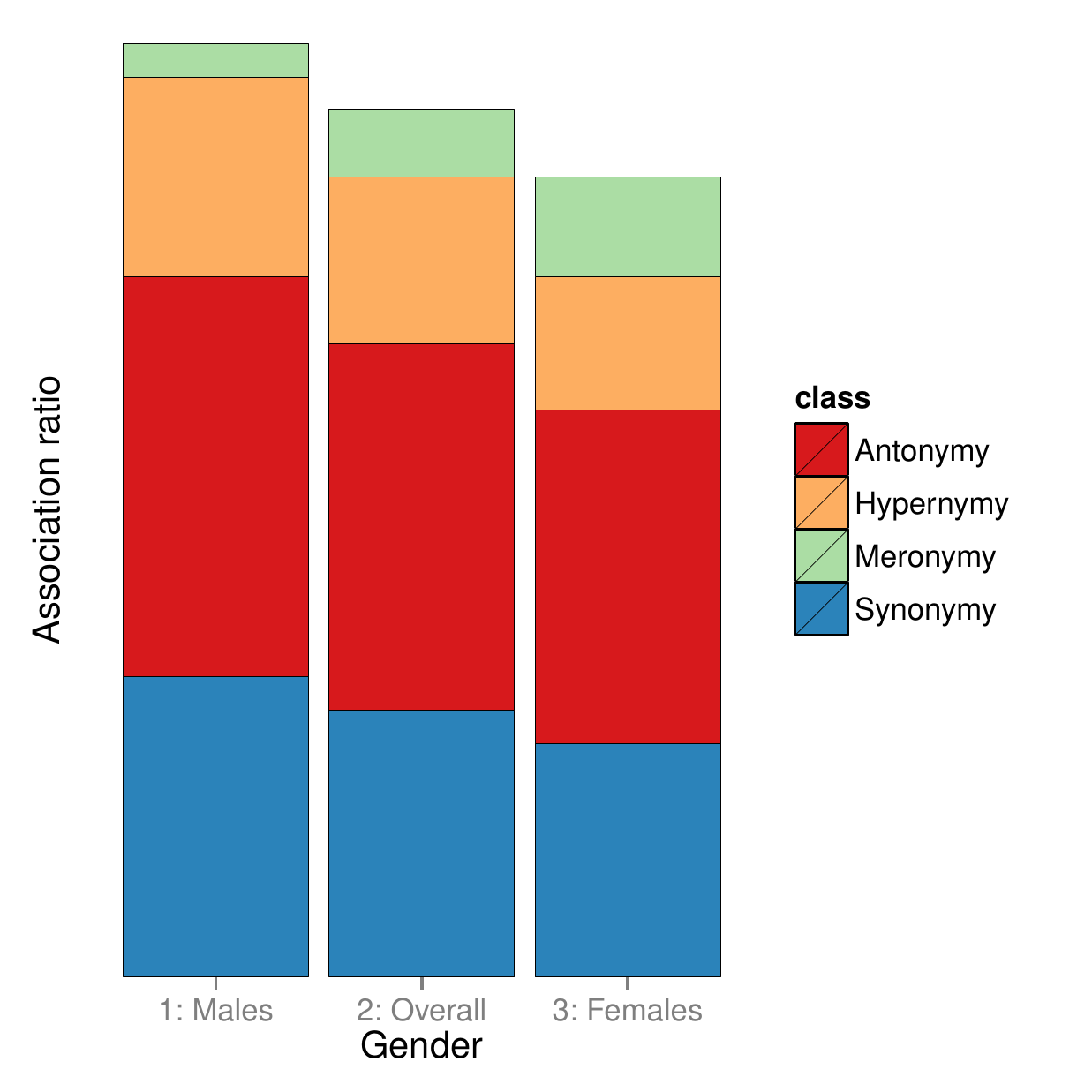}
}{
\caption{Usage of various semantic relations across gender.}
   \label{fig:gen-syn}
}
\capbtabbox{%


 \resizebox{0.55 \textwidth}{!}{ 
  \begin{tabular}{|c|c|c|}
    \hline
    \textit{Stimulus$-$Reaction} & \textit{ Ngram freq. } & \textit{Relations} \\ \hline \hline
     yellow$-$colour & 241 & $-$ \\
     Russia$-$country & 110 & hyponymy\\
     morning$-$good & 445 & $-$ \\
	 mouth$-$face & 6 & part meronymy \\
	 medicine$-$clinic & 0 & domain \\
	 public$-$social & 0 & synonymy \\
	 help$-$find & 33 & $-$ \\
	 most$-$outstanding & 27 & $-$ \\
	 ask$-$answer & 0 & antonymy \\
	 here$-$there & 18 & antonymy \\
	 write$-$letter & 218 & $-$ \\
	 impression$-$emotion & 0 & hyponymy \\
     \hline
  \end{tabular}
  
}{
\caption{An example list of associative pairs and corresponding n-grams along with relations.}
      } 
  \label{tab: ex_ass}
}

\end{floatrow}
\end{figure}

\section{Experiments}
\subsection{Association types analysis}
Aiming to observe possible differences in distribution of association patterns among categories of respondents, we match associations against two major patterns as follows. 

First, we check whether a stimulus $-$ response pair matches an ngram observed in text corpora. We measure a smoothed sum of matched ngram frequencies:
\begin{equation}
	S = \sum\limits_{a \in A}\left\lbrace\begin{array}{ll}\log f(a), &\mbox{ iff } f > 0,\\0,& \mbox{ iff } f = 0 \end{array} \right\rbrace
\end{equation}
where $a$ is an association pair, $A$ is a set of all association pairs, $f(a)$ is a corpus frequency of an ngram produced of $a$ association.

We consider bigrams for single-word responses (the vast majority of cases). If we have two-word response, we match it against trigrams. Responses containing more than two words are treated as non-matching any known ngrams. We tried to match each association both in forward (\textit{stimulus$\rightarrow$response}) and backward (\textit{response$\rightarrow$stimulus}) direction, and each side (\textit{stimulus} and \textit{response}) was supplied both as is and in a lemmatized form.\footnote{We used \textit{mystem}\cite{segalovich2003fast} to extract lemmas.} By doing that, we actually match each association against eight candidate ngrams, and we pick the maximum frequency observed over those ngrams. We used National Corpus of Russian Language  \footnote{http://www.ruscorpora.ru/corpora-freq.html} as source for ngram frequencies.

\begin{figure}
\begin{floatrow}[2]
\ffigbox[0.45\textwidth]{%
\includegraphics[scale=0.27]{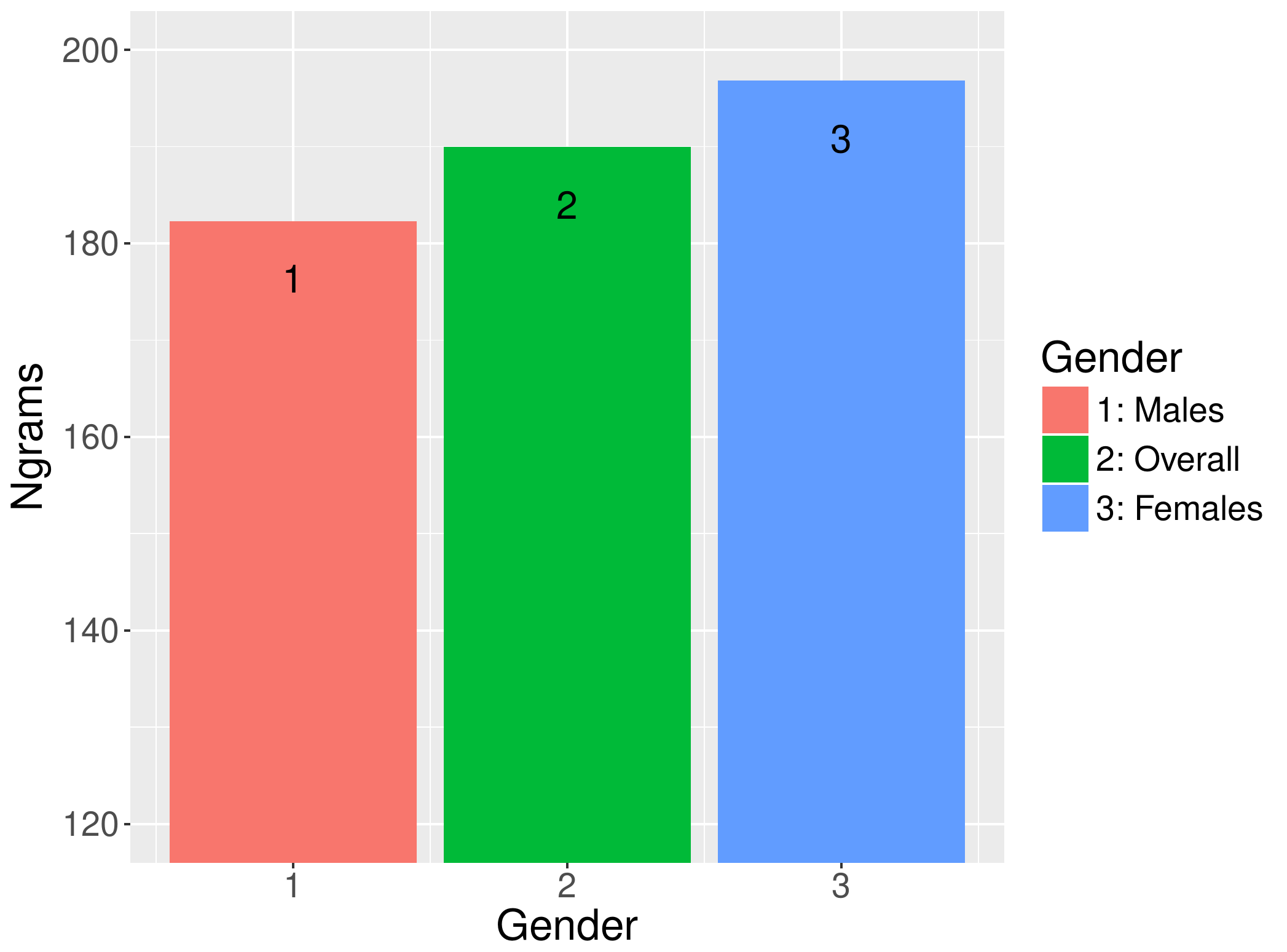}
}{%
	\caption{Usage of ngrams by males and females.}
   \label{fig:ngrams-gen}
}
\capbtabbox{%
  \begin{tabular}{l | l |l }
    \textbf{Top$-$10} & \textbf{ Karaulov's} & \textbf{ Sharoff's} \\ \hline\hline
    Human being & Human being & Year \\ \hline
    Money & Home/House & Human being \\ \hline
    Friend & Money & Time \\ \hline
    Home/House & Day & Business \\ \hline
    Life & Friend & Life \\ \hline
    Day & Home  & Day \\ \hline
    World/Peace & Male & Hand \\ \hline
    Big/Large & Fool & Work \\ \hline
    Time & Business & Word \\ \hline
    Child & Life & Place \\ \hline
  \end{tabular}
}{%
  \caption{A top$-$10 list of the most frequent reactions received in the current study and Karaulov's experiment compared to the top frequent words from Sharoff's list.}
  \label{tab:top-10}
}
\end{floatrow}
\end{figure}

Second, we extract associations that correspond to basic thesaurus relations such as synonyms, antonyms, hypernyms/hyponyms, meronyms/holonyms, and cause/effect. We use Russian WordNet\footnote{http://wordnet.ru} \cite{loukachevitch2016creating} as our initial data source, where we count the number of matching associations for every relation type. Table 1 presents a list of associations and corresponding extracted ngram frequencies and relations.

We measure the values listed above as percentages to the total number of responses. We have done it for the full dataset and also for slices selected by respondent's gender or specialization.

\begin{figure}
\includegraphics[scale=0.43]{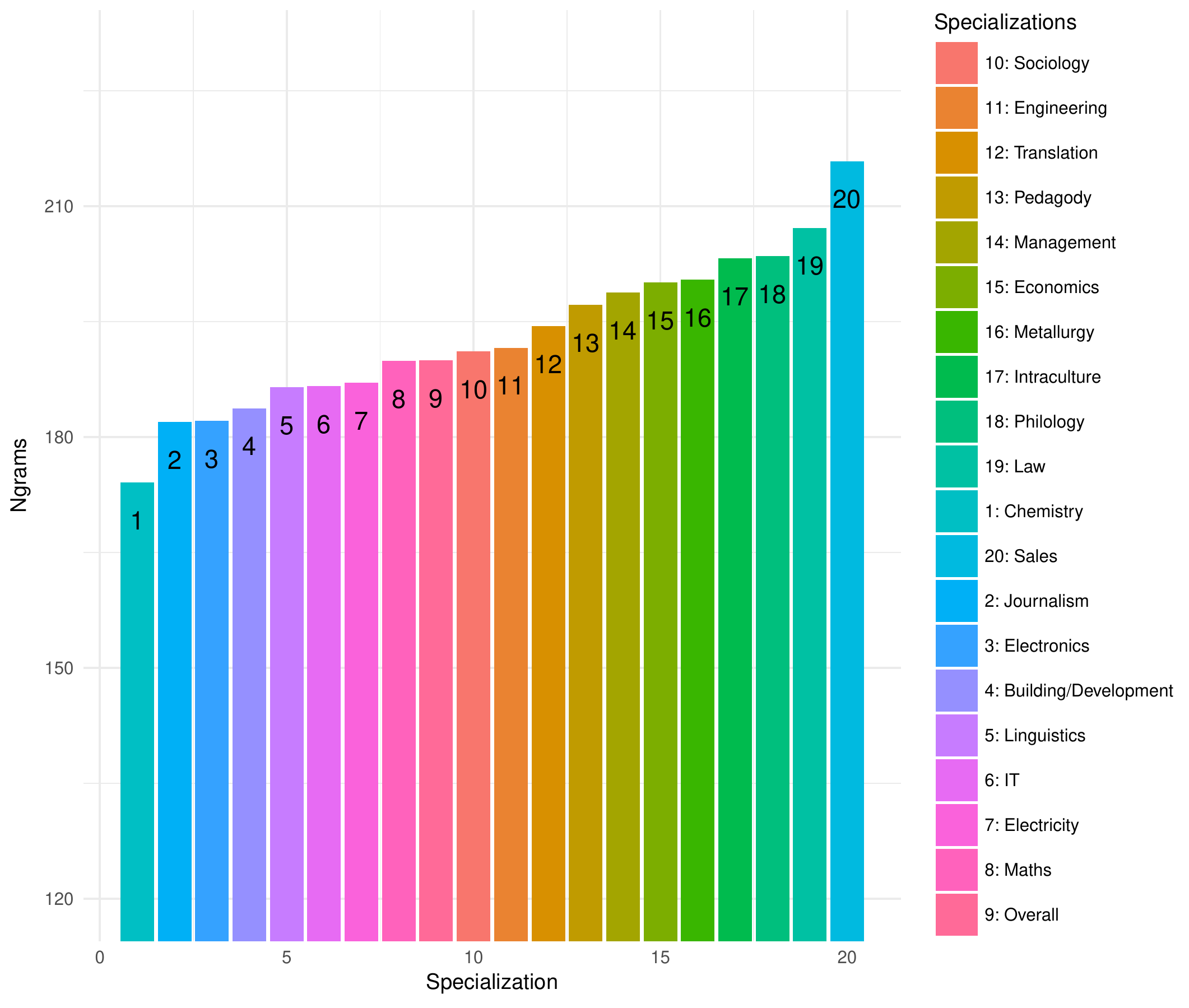}
\caption{Usage of Ngrams across various specializations.}
   \label{fig:ngrams}
\end{figure}

\begin{figure}
\includegraphics[scale=0.43]{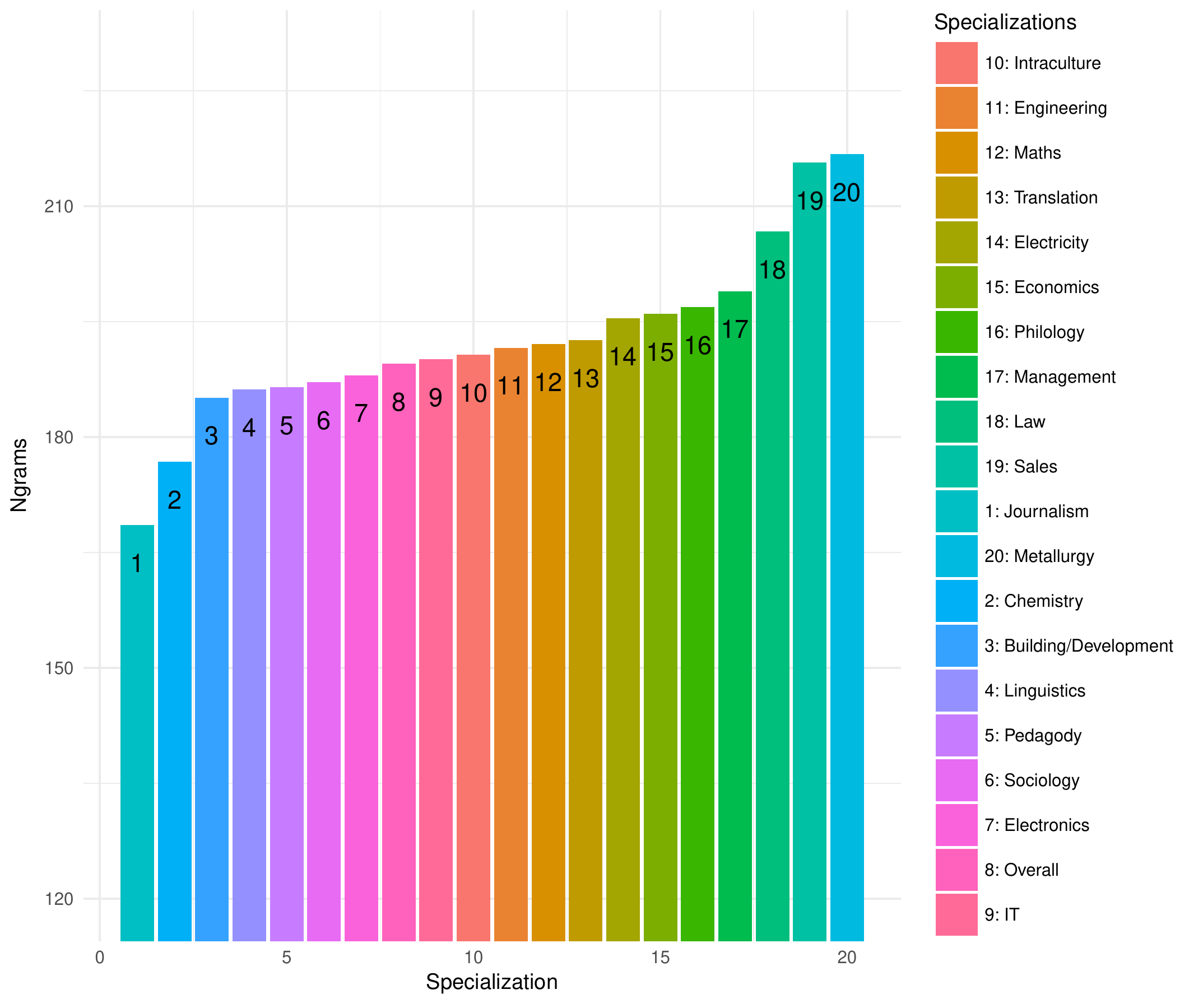}
\caption{Usage of Ngrams across various specializations. Normalised over the gender.}
   \label{fig:ngrams-norm}
\end{figure}

\begin{figure}
\includegraphics[width=0.7\textwidth]{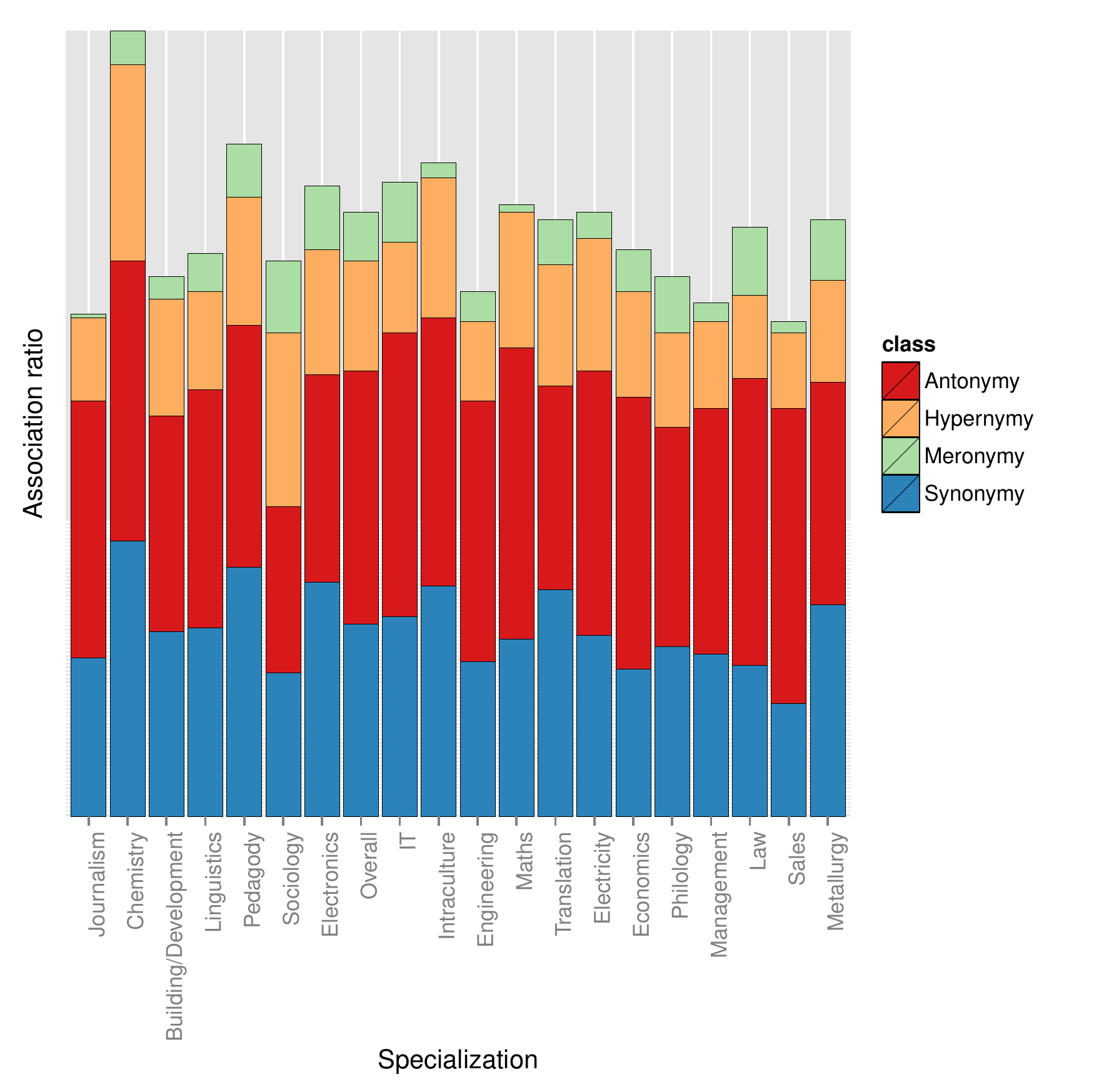}
\caption{Usage of various semantic relations across various specializations.}
\label{fig:synsets}
\end{figure}

Figures \ref{fig:ngrams-gen} and \ref{fig:gen-syn} show that men are more biased towards using semantically inspired associations (paradigmatic) whereas women are more likely to produce ngrams (syntagmatic)\footnote{With p-value$<0.001$}. We observe a similar pattern (i.e. syntagmatic inversely related to paradigmatic) by looking at the specializations. Figure \ref{fig:ngrams} presents $S$ values for the top-20 most popular specializations. For instance, ``chemistry'' presents the highest scores in semantic relations whereas lower than average for ngrams. On the other hand, in the case of ``sales'' it is completely opposite. Note that most of the technical specializations and natural sciences demonstrate high scores for paradigmatic association types. We supposed that this is due to correlation between gender and occupation and the fact that gender still plays a significant role in the process of choosing the future career. In order to test that hypothesis, we calculated ngram usage figures normalized over gender (Fig.~\ref{fig:ngrams-norm}). In this context, a normalized value is a half-sum of two corresponding average values, each one being computed over its respective respondent gender. The normalization didn't seem to smooth differences in ngram usage over various specializations. Therefore, one may conclude that the specialization standalone plays as a significant factor influencing word association patterns.   

\subsection{Personalization of vector models}
Now we turn to our experiments with associative vector models. We propose gender and specialization specific associative models. In order to create the models, we first slice the data by the proposed attribute. For instance, for ``gender'' we take two subsets corresponding to ``male'' and ``female'', respectively.\footnote{The dataset is quite balanced and we have roughly the same number of questionnaires for both male and female participants.} As we described earlier, their association frequency distributions present some differences which we would like our model to capture.
Unlike traditional language modelling task, here we only rely on stimulus-reaction pair frequencies. Therefore, we consider SVD-PPMI\cite{levy2015improving} approach to get the distributed vector representations. The method had been shown to perform on par with neural models\cite{levy2015improving,vylomova2015take}, such as word2vec\cite{Mikolov+:2013b}. It is also less expensive in terms of the time complexity and better fits our task setting.\footnote{We used the model implementation from \url{https://bitbucket.org/omerlevy/hyperwords}. We set the size of the context window to 1 (left and right words), embedding size to 100, context distribution smoothing of 0.75, token threshold value of 5, all the other parameters were left with their default values.}  We additionally train a baseline model on the full dataset to compare it to the personalised models. We would like to emphasize that usage of distributed vector models allows us to go beyond the scope of direct associations and generalize better.

Table \ref{tab: top10-nn} presents several examples for the top 10 nearest neighbours of ``male'', ``female'' and baseline models. In this case, we observe mainly semantic differences. Notice a substantial variation in the predictions of the models if we provide them with ``I'' stimulus. Table \ref{tab: top5-nn} illustrates the models work for ``sales'' and ``publishing'' occupation types. In general, we find the ability to provide gender- and specialization-sensitive information useful for addressing the issues related to social language variation. 

Table \ref{tab: top5-nn2} additionally provides the difference in the model's predictions for two  locations in Chelyabinsk oblast: a small town of Asha and an industrial city of Magnitogorsk.

There is no consensus in the research community on how to evaluate the associative models. Most of the methods are based on direct comparison of statistical characteristics of the distributions of reactions each of the models generates. Moreover, to our knowledge, no theoretical framework or quality assessment or measures had been proposed for that so far. Therefore, we consider this part of research for our future studies.

\begin{table}[!h]
\centering
  \begin{adjustbox}{width=1\columnwidth}

  \begin{tabular}{|c|c|c|c|c|c|c|c|c|}
    \hline
      \multicolumn{3}{|c|}{\textbf{Effectiveness}} &
       \multicolumn{3}{|c|}{\textbf{I}} & 
       \multicolumn{3}{|c|}{\textbf{Work/Job}} \\
    \textit{All} & \textit{ Male } & \textit{ Female} & \textit{All} & \textit{ Male } & \textit{ Female} & \textit{All} & \textit{ Male } & \textit{ Female} \\ \hline \hline
     result & usefulness & result & don't think & workaholic & ego & labour & labour & labour \\
     practicality & result & process  & stupid & bummer & you & well-paid & high-paid & effort\\

     diligence & diligence & diligence & ego & lazy & individual & stock exchange & stock exchange & diligence \\
     usefulness & practicality & quality & individual & quitter & he & deal & to work  & ennoble \\
     perspective & labour & science & nihilist & idler & we & ennoble & worker & deal \\
     process & quality & perspective & Alex & pronoun & selfishness & succeed & deal & hard \\
     quality & high paid & aspiration & Narcissus & student & everyone & effort & activity & worthy\\
     stability & work/job & practicality & loser & sloven & myself & hard-working & office & workaholic \\
     labour & utility & usefulness & nothingness & loafer & Narcissus & diligence & effort & salary \\
     ambition & absolute & progress & selfish & sluggard & selfishness & to work & diligence & fervor \\
    \hline
  \end{tabular}
    \end{adjustbox}
      \caption{A top$-$10 list of the nearest neighbours for each of the models.}
  \label{tab: top10-nn}
\end{table}

\begin{table}[!h]
\centering
  \begin{adjustbox}{width=1\columnwidth}
\scriptsize
  \begin{tabular}{|c|c|c|c|c|c|c|c|c|}
    \hline
      \multicolumn{3}{|c|}{\textbf{Time}} &
       \multicolumn{3}{|c|}{\textbf{Money}} & 
       \multicolumn{3}{|c|}{\textbf{Red}} \\
    \textit{All} & \textit{ Sales } & \textit{ Publishing} & \textit{All} & \textit{ Sales } & \textit{Publishing} & \textit{All} & \textit{ Sales } & \textit{Publishing} \\ \hline \hline
     second & together & last & spend & gold & income & October & color & anger \\
     60 & result & minute  & deficit & a lot of & fuel & square & sun & Lenin\\
     minute & past & deficit & coin & wealth & import & orange & lamp & edge \\
     hour & timely & class & cash & give & gas  & Lenin &yellow  & spite \\
     clock & evening & long & tax & rich & penny & revolution & blue & revolution \\
     \hline
  \end{tabular}
    \end{adjustbox}
      \caption{A top$-$5 list of the nearest neighbours for ``Sales'' and ``Publishing'' specializations.}
  \label{tab: top5-nn}
\end{table}

\begin{table}[!h]
\centering
  \begin{adjustbox}{width=0.9\columnwidth}
\scriptsize
  \begin{tabular}{|c|c|c|c|c|c|}
    \hline
      \multicolumn{2}{|c|}{\textbf{Time}} &
       \multicolumn{2}{|c|}{\textbf{Money}} & 
       \multicolumn{2}{|c|}{\textbf{Red}} \\
     \textit{ Asha } & \textit{ Magnitogorsk} & \textit{ Asha } & \textit{Magnitogorsk} & \textit{ Asha } & \textit{Magnitogorsk} \\ \hline \hline
     of fame & boring & expensive & no money & colour & skin colour \\
     second & waiting & to give & numbers    & bright & green \\
     clock & clock    & salary & hope        & white & yellow \\
     no time & minute & income & debt        & black  & colour \\
     back & train     & to sell & change    & blue & traffic lights\\
     \hline
  \end{tabular}
    \end{adjustbox}
      \caption{A top$-$5 list of the nearest neighbours for ``Asha'' and ``Magnitogorsk'' cities.}
  \label{tab: top5-nn2}
\end{table}

\section{Conclusion}
We presented a new dataset for Russian verbal associations. We also showed that social factors such as gender and specialization provide a significant amount of information on the type of association.
Finally, we also proposed a gender-sensitive associative model and demonstrated the significance of incorporating of social factors into the traditional NLP models. 

\section{Acknowledgments}
We would like to thank all reviewers for their valuable comments and suggestions for future research directions. The first author was supported by the Melbourne International Research Scholarship (MIRS). 

%
%
%




\bibliography{eacl2017}

\end{document}